**SERIEs**
Journal of the
Spanish Economic Association

CrossMark

ORIGINAL ARTICLE

# Modelling cross-dependencies between Spain's regional tourism markets with an extension of the Gaussian process regression model

Oscar Claveria[1,4] · Enric Monte[2] ·
Salvador Torra[3]

**Abstract** This study presents an extension of the Gaussian process regression model for multiple-input multiple-output forecasting. This approach allows modelling the cross-dependencies between a given set of input variables and generating a vectorial prediction. Making use of the existing correlations in international tourism demand to all seventeen regions of Spain, the performance of the proposed model is assessed in a multiple-step-ahead forecasting comparison. The results of the experiment in a multivariate setting show that the Gaussian process regression model significantly improves the forecasting accuracy of a multi-layer perceptron neural network used as a benchmark. The results reveal that incorporating the connections between different markets in the modelling process may prove very useful to refine predictions at a regional level.

✉ Oscar Claveria
oclaveria@ub.edu

Enric Monte
enric.monte@upc.edu

Salvador Torra
storra@ub.edu

1   AQR-IREA (Institute of Applied Economics Research), University of Barcelona, Diagonal, 690, 08034 Barcelona, Spain

2   Department of Signal Theory and Communications, Polytechnic University of Catalunya, Jordi Girona, 1-3, 08034 Barcelona, Spain

3   Department of Econometrics and Statistics, Riskcenter-IREA , University of Barcelona, Diagonal, 690, 08034 Barcelona, Spain

4   Department of Econometrics, University of Barcelona, Diagonal 690, 08034 Barcelona, Spain

## 1 Introduction

In recent years there has been a growing interest in machine learning (ML) techniques for economic forecasting (Weron 2014; Gharleghi et al. 2014; Kock and Teräsvirta 2014; Ben Taieb et al. 2012; Crone et al. 2011; Andrawis et al. 2011; Carbonneau et al. 2008). ML is based on the construction of algorithms that learn through experience. The main ML forecasting methods are support vector regression (SVR) and artificial neural network (ANN) models. Plakandaras et al. (2015) propose a hybrid forecasting methodology that combines an ensemble empirical mode decomposition algorithm with a SVR model to forecast the US real house price index. Lin et al. (2012) also combine an algorithm for time series decomposition with a SVR model for foreign exchange rate forecasting. Kao et al. (2013) and Kim (2003) use different SVR models for stock index forecasting. Tay and Kao (2001, 2002) apply support vector machines in financial time series forecasting.

Stasinakis et al. (2015) use a radial basis function ANN to forecast US unemployment. Feng and Zhang (2014) and Aminian et al. (2006) use ANN models in forecasting of economic growth. Sermpinis et al. (2012) and Lisi and Schiavo (1999) make exchange rates predictions by means of several ANNs. Sarlin and Marghescu (2011) generate visual predictions of currency crisis by means of a self-organizing map ANN model. Adya and Collopy (1998) evaluate the effectiveness of ANN models at forecasting and prediction. A complete summary on the use of ANNs with forecasting purposes can be found in Zhang et al. (1998).

Whilst SVR and ANN models have been widely used in economic modelling and forecasting, other ML techniques such as Gaussian process regression (GPR) have been barely applied for forecasting purposes (Andrawis et al. 2011; Ahmed et al. 2010; Banerjee et al. 2008; Chapados and Bengio 2007; Brahim-Belhouari and Bermak 2004; Girard et al. 2003). GPR was originally devised for interpolation. The works of Smola and Bartlett (2001), MacKay (2003), and Williams and Rasmussen (2006) have been key in the development of GPR models. By expressing the model in a Bayesian framework, the authors extend GPR applications beyond spatial interpolation to regression problems. GPR models are supervised learning methods based on a generalized linear regression that locally estimates forecasts by the combination of values in a kernel (Rasmussen 1996). Thus, GPR models can be regarded as a nonparametric tool for regression in high dimensional spaces. One of the limitations of the current methods for GPR is that the framework is inherently one dimensional, i.e. the framework is designed for multiple inputs and a single output. GPR models present one fundamental advantage over other ML techniques: they provide full probabilistic predictive distributions, including estimations of the uncertainty of the predictions. These features make GPR an ideal tool for forecasting purposes.

This paper presents an extension of the GPR model for MIMO forecasting. This approach allows to preserve the stochastic properties of the training series in multiple-step ahead prediction (Ben Taieb et al. 2010). By extending conventional local modelling approaches we are able to model the cross-dependencies between a given set of time series, returning a vectorial forecast. The structure of the proposed model, consists of a batch of univariate forecasting modules based on Gaussian regression, followed by a linear regression that takes into account the cross-influences between the different forecast.

ML methods are particularly suitable to model phenomena that presents nonlinear interactions between the input and the output. The complex nature behind the data generating process of economic variables such as tourism demand, explains the increasing use of ML methods in this area. There is wide evidence in favour of ML methods when compared to time series models for tourism demand forecasting (Akin 2015; Claveria and Torra 2014; Wu et al. 2012, Hong et al. 2011; Chen and Wang 2007; Giordano et al. 2007; Cho 2003; Law 2000 and Law and Au 1999). Tsaur and Kuo (2011) and Yu and Schwartz (2006) use fuzzy time series models to predict tourism demand. Celotto et al. (2012) and Goh et al. (2008) apply rough sets algorithms. Other authors combine different ML techniques in order to refine forecasts of tourism demand (Cang 2014a; Cang and Yu 2014b; Pai et al. 2014; Shahrabi et al. 2013). Peng et al. (2014) use a meta-analysis to examine the relationships between the accuracy of different forecasting models and the data characteristics in tourism forecasting studies. Athanasopoulos et al. (2011) carry a thorough evaluation of various methods for forecasting tourism data.

In spite of the desirable properties of GPR models, there is only one previous study that uses GPR for tourism demand forecasting (Wu et al. 2012). The authors use a sparse GPR model to predict tourism demand to Hong Kong and find that its forecasting capability outperforms those of the autoregressive moving average (ARMA) and SVR models. We attempt to cover this deficit, and to break new ground by proposing an extension of the GPR model for MIMO modelling, and assessing its forecasting performance. We make use of international tourist arrivals to all seventeen regions of Spain. By incorporating the connections in tourism demand to all regions, we generate forecasts to all markets simultaneously. We finally compare the forecasting performance of the GPR model to that of a multi-layer perceptron (MLP) ANN in a MIMO setting. This strategy is cost-effective in computational terms, and seems particularly indicated for regional forecasting.

Several regional studies have been published in recent years (Lehmann and Wohlrabe 2013), but only a few regarding tourism demand forecasting. Gil-Alana et al. (2008) use different time-series models to models international monthly arrivals in the Canary Islands. Bermúdez et al. (2009) generate prediction intervals for hotel occupancy in three provinces of Spain by means of a multivariate exponential smoothing model. The first attempt to use ML methods for tourism demand forecasting in Spain is that of Palmer et al. (2006), who design a MLP ANN to forecast tourism expenditure in the Balearic Islands. Medeiros et al. (2008) develop an ANN-GARCH model to estimate demand for international tourism also in the Balearic Islands. Claveria et al. (2015) compare the forecasting performance of three ANN architectures to forecast tourist arrivals to Catalonia.

The main aim of this study is to provide researchers with a novel approach for MIMO forecasting, and a method for modelling cross-dependencies. The proposed extension of the GPR model to the MIMO framework allows incorporating the relationships between the different response variables in order to generate a vector of predictions.

The study is organized as follows. The next section presents the proposed extension of the GPR model to the MIMO case. In Sect. 3 we briefly describe the data. Section 4 reports the results of the multiple-step ahead forecasting comparison carried out to test the effectiveness of the model. The last section provides a summary of the theoretical and practical implications, and potential lines for future research.

## 2 Methodology: forecasting models

### 2.1 Gaussian process regression (GPR)

GPR was conceived as a method for multivalued interpolation, and was first developed by Matheron (1973) based on the geostatistic works of Krige (1951). The works of MacKay (2003), Williams and Rasmussen (2006) and Smola and Bartlett (2001) have been crucial in the development of GPR. By expressing the model in a Bayesian framework, different statistical methods can be implemented in GPR models. Therefore GPR applications can be extended beyond spatial interpolation to regression problems, estimating the weights of observed values form temporal lags and spatial points using the known covariance structures. Detailed information about GPR can be found in Williams and Rasmussen (2006).

The GPR model assumes that the inputs $x_i$ have a joint multivariate Gaussian distribution characterized by an analytical model of the structure of the covariance matrix (Rasmussen 1996). The key point of the GPR is the possibility of specifying the functional form of the covariance functions, which allows to introduce prior knowledge about the problem into the model. Note that the functional dependency between variables in the covariance function does not need to be a cross product, but can be any function that takes into account the similarity between the input data points and also complies with the properties of a covariance.

An important point in which GPR differs from linear regression, is that the method assumes a probability distribution over the set of functions to be estimated, which allows for determining families of regression functions with specific functional forms. Formally, the training set $D = \{(x_1, y_1), (x_2, y_2), \ldots, (x_n, y_n)\}$ consists of a set of tuples, and it is assumed to be drawn from the following process:

$$y_i = f(x_i) + \varepsilon, \quad \text{with } \varepsilon \sim N(0, \sigma^2), \tag{1}$$

being $x_i$ an input vector in an Euclidean space of dimension $d$, i.e. $R^d$; and $y_i$ the target, which is a scalar output in $R^1$. This framework allows to estimate a function from $R^d \rightarrow R^1$. For notational convenience, we aggregate the inputs and the outputs into matrix $X = [x_1, x_2, \ldots, x_n]$ and vector $y = [y_1, y_2, \ldots, y_n]$ respectively.

The joint distribution of the variables is the conditional Gaussian distribution, which has the following form:

$$p(y/X) = N(0, K(X, X) + \sigma^2 I), \tag{2}$$

where $I$ is the identity matrix, and $K(X, X)$ the covariance matrix, also referred to as the kernel matrix, with elements $K_{ij}(x_i, x_j)$. The kernel function $k(x, x')$ is a measure of the distance between input vectors. The kernel does not need to be strictly a matrix of cross-products between the input vectors. Kernels may incorporate a distance, or an exponential of a distance.

We try several kernel functions: the rational quadratic covariance function, the gamma exponential covariance function, and a radial basis kernel. We obtain the best performance with a radial basis function, as the other kernels allow for long term interactions between distant points in the input space. See MacKay (2003) and Williams and Rasmussen (2006) for a detailed analysis on kernel selection. Kernel functions should reflect the a priori knowledge about the problem at hand. To obtain local forecasts in the space of inputs, we select a covariance matrix that has a component with the shape of a Gaussian so as to model the interactions between nearby points. Specifically, we use an isotropic Gaussian, which is a Gaussian with a covariance proportional to the identity. We additionally introduce a term to account for non-stationarity in the data that corresponds to a dot product between sample points in the covariance matrix.

Therefore, in this study we make use of a radial basis kernel with a linear trend, which assumes a local continuity of the response variable:

$$K_{ij} = k(x_i, x_j) = v^2 \exp\left(-\frac{(x_i - x_j)^T (x_i - x_j)}{2\lambda^2}\right) + \gamma x_i^T x_j + \kappa, \tag{3}$$

where $v^2$ controls the prior variance, and $\lambda$ is a parameter that controls the rate of decay of the covariance by determining how far away $x_i$ must be from $x_j$ for $f_i$ to be unrelated to $f_j$. Note that the underlying operation is framed in the field of interpolation. The hyperparameters $\{v, \lambda, \gamma, \kappa\}$ are estimated by maximum likelihood in:

$$\log(p(y/x)) = -\frac{1}{2}y^T \left[K(X, X) + \sigma^2 I\right]^{-1} y - \frac{1}{2}\log\left|K(X, X) + \sigma^2 I\right| - \frac{n}{2}\log 2\pi. \tag{4}$$

Given the subscripts of the variables that determine the covariance matrix, $f$ and the predictive outputs $f^*$, by making use of the Bayesian inference, the joint posterior distribution is:

$$p(f, f^*/y) = \frac{p(y/f)p(f, f^*)}{p(y/X)}, \tag{5}$$

$$p(f, f^*) \sim N\left(0, \begin{bmatrix} K_{f,f} K_{f^*,f} \\ K_{f,f^*} K_{f^*,f^*} \end{bmatrix}\right), \tag{6}$$

$$p(y/f) \sim N(f, \sigma^2 I), \tag{7}$$

where $K_{f,f}$ is the covariance matrix of the training data, $K_{f*,f}$ a matrix that gives the mapping of the kernel on the combinations of test and train inputs, and $K_{f*,f*}$ the kernel matrix of the test inputs.

The output given by the GPR consists of a Gaussian predictive distribution $p(f*/y)$ that is characterized by mean $\mu$ and variance $\Sigma$. Therefore, the GPR model specification is given by equations:

$$\mu = K(X^*, X)\left[K(X, X) + \sigma^2 I\right]^{-1} y, \tag{8}$$

$$\Sigma = K(X^*, X^*) - K(X^*, X)\left[K(X, X) + \sigma^2 I\right]^{-1} K(X, X^*). \tag{9}$$

In this research, we use the mean value of the distribution as the predicted value of the GPR. For a given set of inputs $X^* = \left[x_1^*, x_2^*, \ldots, x_n^*\right]$, which optionally could consist of a single observation, we compute the output $f^*$ as $\mu$.

In this study we propose an extension of the GRP model for MIMO modelling, basing this extension on an analogy to radial basis functions. In this analogy, each single GPR gives a prediction of the value of each individual predictor, and a multivariate linear regression combines these outputs into a new output vector. That is, we use a set of univariate predictors followed by a matrix product that takes into account the cross-dependencies of the outputs in order to improve the performance of each single GPR. In this case we have a $R^d \rightarrow R^M$ mapping, where $M$ is the dimension of the output. This extension is applied by following a two-step training:

(i) First, we train and generate supervised forecasts for each time series. That is for each multivariate input, we compute a vector of outputs $f^*$ of the trained GPR.

(ii) In the second step, we estimate a regularized linear regression (Haykin 2008) from a training set that consists of tuples, $D^f = \{(f_1, y_1), (f_2, y_2), \ldots, (f_n, y_n)\}$. The coefficients of the matrix corresponding to this regularized linear regression will be denote as $W^{reg}$. Therefore, the predictions, which we denote as $y^*$, are generated by means of the following expression:

$$y^* = W^{reg} f^*. \tag{10}$$

This procedure will be referred to as MIMO GPR.

## 2.2 Multi-layer perceptron (MLP) artificial neural network (ANN)

Many different NN models have been developed since the 1980s. The most widely used feed-forward topology in tourism demand forecasting is the MLP network (Liang 2014; Teixeira and Fernandes 2012; Lin et al. 2011; Zhang and Qi 2005). In feed-forward networks the information runs only in one direction. MLP networks are supervised neural networks that use a simple perceptron model as a building block. The topology is based on layers of parallel perceptrons, with a nonlinear function at each perceptron. The specification of a MLP network with an input layer, a hidden layer, and an output layer is defined by:

$$y_t = \beta_0 + \sum_{j=1}^{q} \beta_j \, g\left(\sum_{i=1}^{p} w_{ij}x_{t-i} + w_{0j}\right),$$
$$\{x_{t-i}; \; i = 1, \ldots, p\},$$
$$\{w_{ij}; \; i = 1, \ldots, p; \; j = 1, \cdots, q\},$$
$$\{\beta_j; \; j = 1, \ldots, q\},$$

$(11)$

where $y_t$ is the output vector of the network at time $t$; $x_{t-i}$ is the input value at time $t - i$, where $i$ stands for the number of lags that are used to introduce the context of the actual observation; $\beta_j$ are the weights connecting the output of the neuron $j$ at the hidden layer with the output neuron; $w_{ij}$ stand for the weights of neuron $j$ connecting the input with the hidden layer, and $g$ is the nonlinear function of the neurons in the hidden layer. We denote $q$ as the number of neurons in the hidden layer, which determines the MLP network's capacity to approximate a given function. We use values from 5 to 30 with an increase proportional to the length of the forecasting horizon.

As with the GPR model, we also apply a MIMO approach by estimating a regularized linear regression (Haykin 2008), and generate the vectorial forecasts using the set of regularized coefficients.

The estimation of the parameters is done by cross-validation (Bishop 2006). We divide the database into three sets: training, validation and test. The validation set is used to determine the stopping time for the training and the number of neurons in the hidden layer. The test set is used to estimate the generalization performance of the network, that is the performance on unseen data (Bishop 1995; Ripley 1996).

Once the topology of the model is specified, the estimation of the weights of the networks can be done by means of different algorithms. In this study we use the Levenberg–Marquardt (LM) algorithm. To avoid the possibility that the search for the optimum value of the parameters finishes in a local minimum, we use a multi-starting technique that initializes the NN several times for different initial random values, trains the network and chooses the one with the best result on the validation set.

Based on these considerations, the first 96 monthly observations (from January 1999 to December 2006) are selected as the initial training set, the next sixty (from January 2007 to December 2011) as the validation set, and the last 15 % as the test set. For an iterated multi-step-ahead forecasting comparison the partition between train and test sets is done sequentially: as the prediction advances, past forecasts are successively incorporated to the training database in a recursive way.

## 3 Dataset

Tang and Abosedra (2015), Pérez-Rodríguez et al. (2015) and Chou (2013) have shown the important role of tourism in economic growth. In this study we use data on international tourism demand to all regions of Spain provided by the Spanish Statistical Office (National Statistics Institute—INE—http://www.ine.es). Data include the monthly number of tourist arrivals at a regional level over the time period 1999:01 to 2014:03.

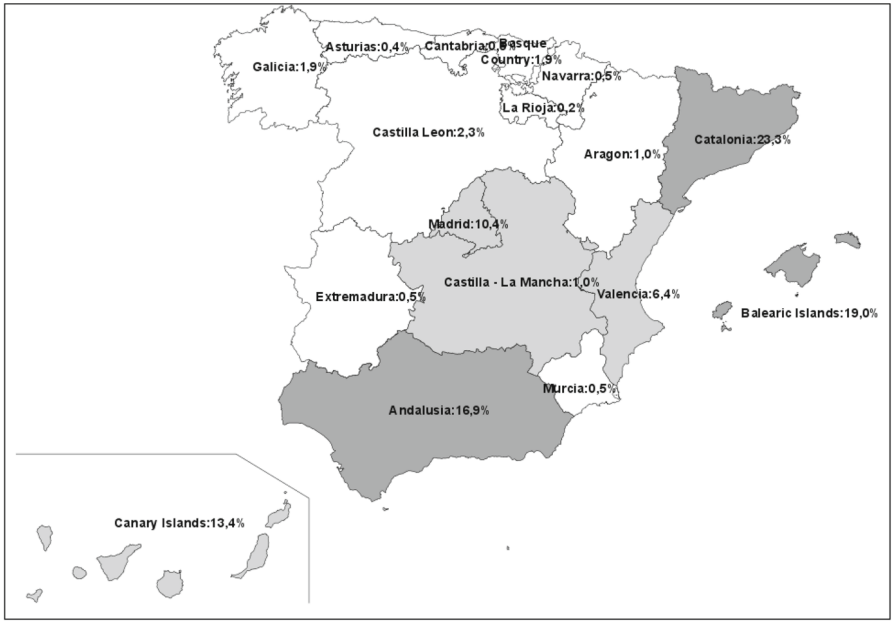

**Fig. 1** Frequency distribution of tourist arrivals to Spain by region (mean from 1999:01 to 2014:03)

Tourism is a key economic sector in Spain. It represents about 11 % of GDP and 12.7 % of total employment (WTTC 2016). Low oil prices and the dollar-euro exchange rate have had a positive impact on the inbound tourism. The country registered a record of 64.9 million international tourist arrivals in 2014, which represented a growth of 7.1 % over the previous year. The main source markets are the United Kingdom, France, Germany and Italy. Tourism is highly concentrated in the summer season in most regions. Tourist arrivals are also concentrated in the main regional destinations: Catalonia, the Balearic Islands, the Canary Islands, and Andalusia.

A MIMO approach to regional economic modelling is particularly appropriate when the desired outputs are connected (Claveria et al. 2015). In Fig. 1 we present the frequency distribution of tourist arrivals by region during the sample period. We can see that most tourist arrivals are concentrated in the Mediterranean coast and the islands, being Catalonia, the Balearic Islands and Andalusia the regions that received the higher number of tourist arrivals, which almost amounted to 60 % of total tourist demand.

Table 1 shows a descriptive analysis of the data for the out-of-sample period (2012:01 to 2014:03). The mean of tourist arrivals shows that the main destinations are Catalonia, the Balearic Islands and Andalusia. The Balearic Islands and Catalonia present the highest peaks in demand. Arrivals to the Balearic Islands show the highest dispersion.

**Table 1** Descriptive analysis of foreign tourist arrivals (2012:01–2014:03)

| Region | Minimum | Maximum | Mean | Standard deviation | Skewness | Kurtosis |
|--------|---------|---------|------|--------------------|----------|----------|
| Andalusia | 237,744 | 770,987 | 496,549.3 | 192,639.5 | −0.13 | −1.70 |
| Aragon | 14,792 | 59,194 | 31,359.3 | 12,915.7 | 0.46 | −0.97 |
| Asturias | 3,347 | 33,714 | 14,092.4 | 9,597.3 | 0.56 | −1.11 |
| Balearic Islands | 23,446 | 1,387,491 | 551,636.0 | 545,838.5 | 0.44 | −1.63 |
| Canary Islands | 385,225 | 619,311 | 499,375.4 | 56,536.1 | 0.07 | −0.35 |
| Cantabria | 3,577 | 32,070 | 14,870.8 | 10,503.4 | 0.42 | −1.49 |
| Castilla-Leon | 23,317 | 134,683 | 67,108.0 | 36,092.0 | 0.25 | −1.51 |
| Castilla-La Mancha | 13,209 | 36,444 | 24,822.6 | 8,266.0 | −0.16 | −1.65 |
| Catalonia | 336,275 | 1,442,017 | 801,443.7 | 369,301.5 | 0.28 | −1.42 |
| Valencia | 103,522 | 322,857 | 207,634.5 | 67,098.1 | −0.10 | −1.40 |
| Extremadura | 6,797 | 24,817 | 14,115.8 | 5,045.3 | 0.15 | −1.00 |
| Galicia | 15,890 | 126,066 | 60,342.3 | 40,727.1 | 0.30 | −1.63 |
| Madrid | 240,349 | 432,430 | 342,618.9 | 62,420.2 | −0.17 | −1.47 |
| Murcia | 8,607 | 22,480 | 15,126.3 | 3,763.6 | 0.11 | −0.90 |
| Navarra | 4,416 | 35,152 | 16,346.1 | 10,355.8 | 0.49 | −1.27 |
| Basque Country | 31,597 | 142,644 | 70,214.0 | 34,130.0 | 0.59 | −0.83 |
| La Rioja | 2,157 | 15,404 | 6,824.1 | 4,190.2 | 0.58 | −0.78 |
| Total | 1,583,237 | 5,283,691 | 3,234,479 | 1,337,386 | 0.17 | −1.69 |

## 4 Results of the experiment

In a recent and comprehensive comparison on the M3 dataset for the major ML models for time series forecasting, Ahmed et al. (2010) find that MLP ANN and GPR models present the best results. Therefore, to assess the forecasting performance of the proposed extension of the GPR model, we compare it to that of a MLP ANN in a MIMO setting. First, we estimate the models and generate forecasts for different forecast horizons (h = 1, 2, 3 and 6 months). Multiple-step ahead forecasts are generated by means of a rolling scheme.

Second, by means of several forecast accuracy measures, we summarize the results for the out-of-sample period. First, we compute the relative mean absolute percentage error (rMAPE) statistic (Table 2), that ponders the MAPE of the model under evaluation against the MAPE of the benchmark model. Next, we run the Diebold and Mariano (DM) test (Diebold and Mariano 1995) using a Newey–West type estimator, and a modified DM (M-DM) test (Harvey et al. 1997) to analyse whether the reductions in MAPE are statistically significant (Table 2). The null hypothesis of the test is that the difference between the two competing series is non-significant. A negative sign of the statistic implies that the MLP ANN model has bigger forecasting errors.

Table 2 shows the overall performance of the compared forecasting models on all regions. There are only two regions in which the ANN model presents lower MAPE

**Table 2** Forecast accuracy: MIMO GPR vs. MIMO ANN model (2013:01–2014:01)

| Region | Statistic | Forecast horizon | | | |
|---|---|---|---|---|---|
| | | h = 1 | h = 2 | h = 3 | h = 6 |
| Andalusia | rMAE | 0.803 | 0.724 | 0.821 | 0.977 |
| | DM | −3.828 | −5.386 | −4.619 | −2.113 |
| | M-DM | −15.312 | −22.948 | −20.894 | −11.242 |
| Aragon | rMAE | 0.805 | 0.781 | 0.825 | 1.219 |
| | DM | −0.376 | −2.204 | −2.294 | 2.192 |
| | M-DM | −1.504 | −9.391 | −10.377 | 11.663 |
| Asturias | rMAE | 0.683 | 0.742 | 0.889 | 0.845 |
| | DM | −1.301 | −2.823 | −2.517 | 0.660 |
| | M-DM | −5.204 | −12.028 | −11.386 | 3.512 |
| Balearic Islands | rMAE | 1.074 | 0.54 | 0.562 | 1.586 |
| | DM | −1.102 | −3.404 | −3.553 | −0.239 |
| | M-DM | −4.408 | −14.504 | −16.072 | −1.272 |
| Canary Islands | rMAE | 1.132 | 1.073 | 1.01 | 0.967 |
| | DM | 2.768 | 0.891 | 0.256 | −0.898 |
| | M-DM | 11.072 | 3.796 | 1.158 | −4.778 |
| Cantabria | rMAE | 0.785 | 0.852 | 0.689 | 0.769 |
| | DM | −2.499 | −3.326 | −3.798 | 0.058 |
| | M-DM | −9.996 | −14.171 | −17.180 | 0.309 |
| Castilla-Leon | rMAE | 0.751 | 0.632 | 0.601 | 0.934 |
| | DM | −1.948 | −4.885 | −3.15 | 1.040 |
| | M-DM | −7.792 | −20.814 | −14.249 | 5.533 |
| Castilla-La Mancha | rMAE | 0.623 | 0.432 | 0.47 | 0.669 |
| | DM | −2.987 | −4.548 | −3.781 | −2.239 |
| | M-DM | −11.948 | −19.378 | −17.103 | −11.913 |
| Catalonia | rMAE | 0.756 | 0.71 | 0.857 | 1.145 |
| | DM | −1.635 | −2.107 | −1.683 | 2.405 |
| | M-DM | −6.540 | −8.977 | −7.613 | 12.796 |
| Valencia | rMAE | 0.902 | 0.942 | 0.835 | 0.981 |
| | DM | −1.341 | −2.429 | −3.153 | −2.744 |
| | M-DM | −5.364 | −10.349 | −14.262 | −14.599 |
| Extremadura | rMAE | 0.915 | 0.919 | 0.954 | 0.861 |
| | DM | −0.685 | −1.863 | −2.259 | −1.933 |
| | M-DM | −2.740 | −7.938 | −10.218 | −10.285 |
| Galicia | rMAE | 0.812 | 0.761 | 0.862 | 0.775 |
| | DM | −1.536 | −3.409 | −2.314 | −0.4 |
| | M-DM | −6.144 | −14.525 | −10.467 | −2.128 |
| Madrid | rMAE | 1.182 | 0.986 | 1.078 | 1.014 |
| | DM | 0.361 | 0.325 | 0.950 | 0.962 |
| | M-DM | 1.444 | 1.385 | 4.297 | 5.118 |

**Table 2** continued

| Region | Statistic | Forecast horizon | | | |
|--------|-----------|--------|--------|--------|--------|
| | | h = 1 | h = 2 | h = 3 | h = 6 |
| Murcia | rMAE | 1.019 | 0.845 | 0.959 | 0.967 |
| | DM | −0.007 | −3.069 | −4.365 | −3.397 |
| | M-DM | −0.028 | −13.076 | −19.745 | −18.074 |
| Navarra | rMAE | 0.735 | 0.643 | 0.925 | 1.205 |
| | DM | −1.395 | −3.052 | −2.534 | 2.11 |
| | M-DM | −5.580 | −13.004 | −11.462 | 11.226 |
| Basque Country | rMAE | 0.838 | 0.801 | 0.808 | 1.074 |
| | DM | −2.142 | −1.76 | −1.416 | 1.008 |
| | M-DM | −8.568 | −7.499 | −6.405 | 5.363 |
| La Rioja | rMAE | 0.951 | 0.653 | 0.6 | 0.908 |
| | DM | −0.533 | −3.585 | −3.221 | −0.046 |
| | M-DM | −2.132 | −15.275 | −14.570 | −0.245 |

The rMAPE ponders the MAPE of the model under evaluation against the MAPE of the benchmark model. We use a MIMO MLP ANN model as a benchmark. The 5 % level critical value for the Diebold–Mariano (DM) loss-differential test statistic for predictive accuracy is 2.028. M-DM refers to the modified DM test statistic values: the Canary Islands and Madrid, but the differences between both methods are not significant in three out of the four forecasting horizons.

In most regions, the best forecasting performance is obtained with the MIMO GPR model. Nevertheless, we find differences across regions. While in the Balearic Islands, Castilla-Leon, Catalonia, Extremadura, the Basque Country and La Rioja, GPRs outperform the ANNs, this improvement is not significant for all forecast horizons, especially for h = 1 and h = 3. As a result, we find that the improvement in forecast accuracy of the MIMO GPR with respect to the MIMO ANN forecasts becomes more prominent for intermediate forecasting horizons (2 and 3 months).

In general, we obtain the most accurate predictions for longer forecast horizons (h = 6). This can be attributed to the fact that ML methods use nonlinear functions that can account for saturation effects. The structure of these models can be empirically estimated, so the interactions between input variables can be learned from the data.

In Fig. 2 we compare the rMAPE results for one- and three-month ahead forecasts (h = 1 and h = 3). The graph indicates that there are only four regions in which the rMAPE is higher than one for h = 1, that is the ANN outperforms the GPR model for one-month ahead forecasts: the Balearic and the Canary Islands, Madrid and Murcia. Of these four regions, just two (the Canary Islands and Madrid) still obtain a rMAPE higher than one for h = 3. The fact that these two regions do not present seasonal patterns, suggests that GPR are more suitable for seasonal forecasting than ANN models.

Finally, in Table 3 we present the results of the percentage of periods with lower absolute error (PLAE) statistic proposed by Claveria et al. (2015). The PLAE can be regarded as a variation of the 'percent better' measure used to compare the forecast

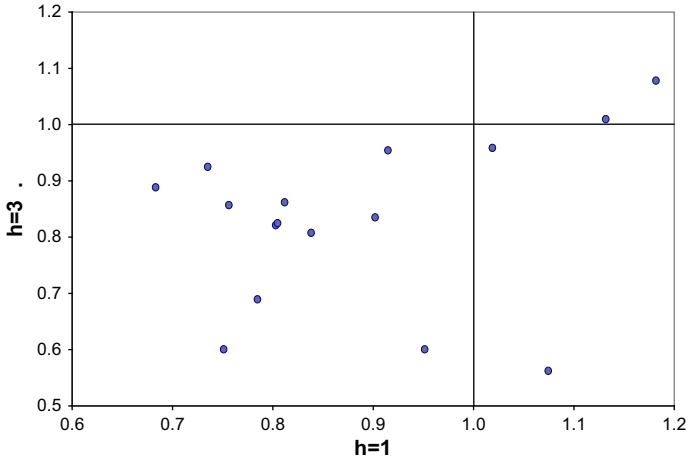

**Fig. 2** Dispersion graph between rMAPFE for h = 1 and h = 3

**Table 3** Forecast accuracy: PLAE: GPR with respect to MLP ANN model (2013:01–2014:01)

| Region | Forecast horizon | | | |
| --- | --- | --- | --- | --- |
| | h = 1 | h = 2 | h = 3 | h = 6 |
| Andalusia | 69.2 | 53.8 | 61.5 | 61.5 |
| Aragon | 38.5 | 53.8 | 69.2 | 15.4 |
| Asturias | 61.5 | 76.9 | 53.8 | 23.1 |
| Balearic Islands | 76.9 | 69.2 | 76.9 | 61.5 |
| Canary Islands | 30.8 | 23.1 | 46.2 | 53.8 |
| Cantabria | 76.9 | 69.2 | 76.9 | 46.2 |
| Castilla-Leon | 69.2 | 69.2 | 76.9 | 46.2 |
| Castilla-La Mancha | 76.9 | 61.5 | 76.9 | 69.2 |
| Catalonia | 38.5 | 69.2 | 53.8 | 7.7 |
| Valencia | 38.5 | 46.2 | 53.8 | 53.8 |
| Extremadura | 53.8 | 53.8 | 61.5 | 69.2 |
| Galicia | 69.2 | 76.9 | 53.8 | 53.8 |
| Madrid | 23.1 | 38.5 | 30.8 | 23.1 |
| Murcia | 30.8 | 61.5 | 53.8 | 61.5 |
| Navarra | 76.9 | 76.9 | 69.2 | 15.4 |
| Basque Country | 53.8 | 61.5 | 53.8 | 30.8 |
| La Rioja | 69.2 | 76.9 | 61.5 | 38.5 |

Percentage of PLAE values. The PLAE ratio measures the number of out-of-sample periods with lower absolute errors than the benchmark model (MLP ANN) accuracy of the models to a random walk (Makridakis and Hibon 2000). See Makridakis et al. (1998) and Witt and Witt (1992) for an appraisal of different forecasting accuracy measures. The PLAE is a dimensionless measure based on the CJ statistic for testing market efficiency (Cowles and Jones 1937). In this study we use the MLP ANN model as a benchmark.

The PLAE allows us to compare the forecasting performance between two competing models. This accuracy measure consists of a ratio that gives the proportion of periods in which the model under evaluation obtains a lower absolute forecasting error than the benchmark model. Let us denote $y_t$ as actual value and $\hat{y}_t$ as forecast at period $t = 1, \ldots, n$. Forecast errors can then be defined as $e_t = y_t - \hat{y}_t$. Given two competing models $A$ and $B$, where $A$ refers to the forecasting model under evaluation and $B$ stands for benchmark model, we can then obtain the proposed statistic as follows:

$$PLAE = \frac{\sum_{t=1}^{n} \lambda_t}{n}, \quad \text{where } \lambda_t = \begin{cases} 1 & \textbf{if} \quad \left| e_{t,A} \right| < \left| e_{t,B} \right|, \\ 0 & \textbf{otherwise}. \end{cases} \tag{12}$$

Table 3 shows that the MIMO GPR outperforms the MIMO MLP ANN model in most cases, especially for 2- and 3-month-ahead forecasts. Special mention should be made of the Canary Islands and the Community of Madrid, where the ANN forecasts provide lower PLAE values. This result can be explained by the fact that they are the only regions that do not show seasonal patterns.

To summarize, we find that the overall forecasting performance improves for longer forecast horizons. This evidence confirms previous research by Teräsvirta et al. (2005), who obtain more accurate forecasts with ANN models at long forecast horizons. This result is indicative that ML techniques are particularly suitable for medium and long term forecasting.

Regarding the different methods, we obtain better predictions with the GPR model than with the ANN. This improvement is more generalized for intermediate forecast horizons. Despite being the first study to apply a MIMO approach for GPR forecasting, our results are in line with those obtained by Wu et al. (2012), who find evidence that a sparse GPR model yields better forecasting results than ARMA and SVR models.

## 5 Concluding remarks and future work

As more accurate predictions become crucial for effective management and policy planning, new forecasting methods provide room for improvement. Machine learning techniques are playing a pivotal role in the refinement of economic predictions. With this objective, we propose an extension of the Gaussian process regression model for multiple-input multiple-output forecasting. This approach allows modelling the cross-dependencies between a given set of input variables and generating a vector prediction.

The main theoretical contribution of this study to the economic literature is the development of a new approach to improve the forecasting accuracy of computational intelligence techniques based on machine learning. The increasing economic importance of the tourism industry worldwide has led to a growing interest in new approaches to tourism modelling and forecasting. Making use of the interdependencies in international tourism demand to all seventeen regions of Spain, we design a multiple-input multiple-output framework that incorporates the existing cross-correlations in tourist arrivals to all markets, and allows to estimate tourism demand to all destinations simultaneously.

We evaluate the performance of the new method by comparing it to a standard neural network in an iterative multiple-step-ahead forecasting comparison. We find that the proposed extension of the Gaussian process regression outperforms the benchmark model in most regions, especially for intermediate forecast horizons.

We obtain the best forecasting results for the longest forecast horizons, suggesting the suitability of machine learning techniques for mid and long term forecasting. As a result, our research reveals the suitability of a multiple-output Gaussian process regression model for regional economic forecasting, and highlights the importance of taking into account the connections between different markets when modelling regional variables with machine learning techniques. The assessment of alternative kernel functions on the forecasting accuracy is a question to be addressed in further research.

**Acknowledgments**  We would like to thank Manuel Bagues and two anonymous referees for their useful comments and suggestions.

# References

Adya M, Collopy F (1998) How effective are neural networks at forecasting and prediction? J Forecast 17(5–6):481–495

Ahmed NK, Atiya AF, El Gayar N, El-Shishiny H (2010) An empirical comparison of machine learning models for time series forecasting. Econ Rev 29(4):594–621

Akin M (2015) A novel approach to model selection in tourism demand modeling. Tour Manag 48:64–72

Andrawis RR, Atiya AF, El-Shishiny H (2011) Forecast combinations of computational intelligence and linear models for the NN5 time series forecasting competition. Int J Forecast 27(2):672–688

Aminian F, Dante E, Aminian M, Waltz D (2006) Forecasting economic data with neural networks. Comput Econ 28(1):71–88

Athanasopoulos G, Hyndman RJ, Song H, Wu DC (2011) The tourism forecasting competition. Int J Forecast 27(2):822–844

Banerjee S, Gelfand AE, Finley AO, Sang H (2008) Gaussian predictive process models for large spatial data sets. J Royal Stat Soc Ser B (Stat Methodol) 70:825–848

Ben Taieb S, Sorjamaa A, Bontempi G (2010) Multiple-output modeling for multi-step-ahead time series forecasting. Neurocomputing 73(10–12):1950–1957

Ben Taieb S, Bontempi G, Atiya AF, Sorjamaa A (2012) A review and comparison of strategies for multiple-step ahead time series forecasting based on the NN5 forecasting competition. Experts Syst Appl 39(7):1950–1957

Bermúdez JD, Corberán-Vallet A, Vercher E (2009) Multivariate exponential smoothing: a Bayesian forecast approach based on simulation. Math Comput Simul 79(4):1761–1769

Bishop CM (1995) Neural networks for pattern recognition. Oxford University Press, Oxford

Bishop CM (2006) Pattern recognition and machine learning. Springer, New York

Brahim-Belhouari S, Bermak A (2004) Gaussian process for nonstationary time series prediction. Comput Stat Data Anal 47(3):705–712

Cang S (2014) A comparative analysis of three types of tourism demand forecasting models: individual, linear combination and non-linear combination. Int J Tour Res 16(5):596–607

Cang S, Yu H (2014) A combination selection algorithm on forecasting. Eur J Oper Res 234(1):127–139

Carbonneau R, Laframboise K, Vahidov R (2008) Application of machine learning techniques for supply chain demand forecasting. Eur J Oper Res 184(2):1140–1154

Celotto E, Ellero A, Ferretti P (2012) Short-medium term tourist services demand forecasting with rough set theory. Proc Econ Finance 3:62–67

Chapados N, Bengio Y (2007) Augmented functional time series representation and forecasting with Gaussian processes. In: Schölkopf B, Platt JC, Hoffman T (eds) Advances in neural information processing systems, vol 19. The MIT Press, Cambridge, pp 457–464

Chen KY, Wang CH (2007) Support vector regression with genetic algorithms in forecasting tourism demand. Tour Manag 28(1):215–226

Cho V (2003) A comparison of three different approaches to tourist arrival forecasting. Tour Manag 24(2):323–330

Chou MC (2013) Does tourism development promote economic growth in transition countries? A panel data analysis. Econ Model 33:226–232

Claveria O, Torra S (2014) Forecasting tourism demand to Catalonia: neural networks vs. time series models. Econ Model 36:220–228

Claveria O, Monte E, Torra S (2015) A new forecasting approach for the hospitality industry. Int J Contemp Hosp Manag 27(6):1520–1538

Cowles A, Jones H (1937) Some a posteriori probabilities in stock market action. Econometrica 5(2):280–294

Crone SF, Hibon M, Nikolopoulos K (2011) Advances in forecasting with neural networks? Empirical evidence from the NN3 competition on time series prediction. Int J Forecast 27(2):635–660

Diebold FX, Mariano RS (1995) Comparing predictive accuracy. J Bus Econ Stat 13(2):253–263

Feng L, Zhang J (2014) Application of artificial neural networks in tendency forecasting of economic growth. Econ Model 40:76–80

Gharleghi B, Shaari AH, Shafighi N (2014) Predicting exchange rates using a novel "cointegration based neuro-fuzzy system". Int Econ 137:88–103

Giordano F, La Rocca M, Perna C (2007) Forecasting nonlinear time series with neural network sieve bootstrap. Comput Stat Data Anal 51(7):3871–3884

Girard A, Rasmussen C, Quiñonero-Candela J, Murray-Smith R (2003) Multiple-step ahead prediction for non linear dynamic systems—a Gaussian process treatment with propagation of the uncertainty. In: Becker S, Thrun S, Obermayer K (eds) Advances in neural information processing systems, vol 15. The MIT Press, Cambridge

Gil-Alana LA, Cunado J, De Gracia FP (2008) Tourism in the Canary Islands: forecasting using several seasonal time series models. J Forecast 27:621–636

Goh C, Law R, Mok HM (2008) Analyzing and forecasting tourism demand: a rough sets approach. J Travel Res 46(2):327–338

Harvey DI, Leybourne SJ, Newbold P (1997) Testing the equality of prediction mean squared errors. Int J Forecast 13(1):281–291

Haykin S (2008) Neural networks and learning machines. Prentice Hall, New Jersey

Hong W, Dong Y, Chen L, Wei S (2011) SVR with hybrid chaotic genetic algorithms for tourism demand forecasting. Appl Soft Comput 11(1):1881–1890

Kao LJ, Chiu CC, Lu CJ, Chang CH (2013) A hybrid approach by integrating wavelet-based feature extraction with MARS and SVR for stock index forecasting. Decis Support Syst 54(2):1228–1244

Kim H (2003) Financial time series forecasting using support vector machines. Neurocomputing 55(1–2):307–319

Kock AB, Teräsvirta T (2014) Forecasting performances of three automated modelling techniques during the economic crisis 2007–2009. Int J Forecast 30(2):616–631

Krige DG (1951) A statistical approach to some basic mine valuation problems on the Witwatersrand. J Chem Metall Min Soc S Afr 52(5):119–139

Law R (2000) Back-propagation learning in improving the accuracy of neural network-based tourism demand forecasting. Tour Manag 21(3):331–340

Law R, Au N (1999) A neural network model to forecast Japanese demand for travel to Hong Kong. Tour Manag 20(1):89–97

Lehmann R, Wohlrabe K (2013) Forecasting GDP at the regional level with many predictors. German Econ Rev 16(1):226–254

Liang YH (2014) Forecasting models for Taiwanese tourism demand after allowance for mainland China tourists visiting Taiwan. Comput Indus Eng 74:111–119

Lin C, Chiu S, Lin T (2012) Empirical mode decomposition-based least squares support vector regression for foreign exchange rate forecasting. Econ Model 40:76–80

Lin C, Chen H, Lee T (2011) Forecasting tourism demand using time series, artificial neural networks and multivariate adaptive regression splines: evidence from Taiwan. Int J Bus Adm 2(1):14–24

Lisi F, Schiavo RA (1999) A comparison between neural networks and chaotic models for exchange rate prediction. Comput Stat Data Anal 30(1):87–102

MacKay DJC (2003) Information theory, inference, and learning algorithms. Cambridge University Press, Cambridge

Makridakis S, Wheelwright S, Hyndman R (1998) Forecasting methods and applications, 3rd edn. Wiley, New York

Makridakis S, Hibon M (2000) The M3-competition: results, conclusions and implications. Int J Forecast 16(3):451–476

Matheron G (1973) The intrinsic random functions and their applications. Adv Appl Probab 5(2):439–468

Medeiros MC, McAleer M, Slottje D, Ramos V, Rey-Maquieira J (2008) An alternative approach to estimating demand: neural network regression with conditional volatility for high frequency air passenger arrivals. J Econ 147(1):372–383

Pai P, Hung K, Lin K (2014) Tourism demand forecasting using novel hybrid system. Expert Syst Appl 41(7):3691–3702

Palmer A, Montaño JJ, Sesé A (2006) Designing an artificial neural network for forecasting tourism time-series. Tour Manag 27(4):781–790

Peng B, Song H, Crouch GI (2014) A meta-analysis of international tourism demand forecasting and implications for practice. Tour Manag 45(1):181–193

Pérez-Rodríguez JV, Ledesma-Rodríguez F, Santana-Gallego M (2015) Testing dependence between GDP and tourism's growth rates. Tour Manag 48:268–282

Plakandaras V, Gupta R, Gogas P, Papadimitriou T (2015) Forecasting the US real house price index. Econ Model 45:259–267

Rasmussen CE (1996) The infinite Gaussian mixture model. Adv Neural Inform Process Syst 8:514–520

Ripley BD (1996) Pattern recognition and neural networks. Cambridge University Press, Cambridge

Sarlin P, Marghescu D (2011) Visual predictions of currency crises using self-organizing maps. Intell Syst Acc Finance Manag 18:15–38

Sermpinis G, Dunis C, Laws J, Stasinakis C (2012) Forecasting and trading the EUR/USD exchange rate with stochastic neural network combination and time-varying leverage. Decis Support Syst 54:316–329

Shahrabi J, Hadavandi E, Asadi S (2013) Developing a hybrid intelligent model for forecasting problems: case study of tourism demand time series. Knowl Based Syst 43:112–122

Smola AJ, Bartlett PL (2001) Sparse greedy Gaussian process regression. Adv Neural Inform Process Syst 13:619–625

Stasinakis C, Sermpinis G, Theofilaos K, Karathanasopoulos A (2015) Forecasting US unemployment with radial basis neural networks, Kalman filters and support vector regressions. Comput Econ 1:1–19

Tang CF, Abosedra S (2015) Tourism and growth in Lebanon: new evidence from bootstrap simulation and rolling causality approaches. Empir Econ (forthcoming)

Tay FEH, Kao L (2001) Application of support vector machines in financial time series forecasting. Omega 29(3):309–317

Tay FEH, Kao L (2002) Modified support vector machines in financial time series forecasting. Neurocomputing 48(1–4):847–861

Teixeira JP, Fernandes PO (2012) Tourism time series forecast—different ANN architectures with time index input. Proc Technol 5:445–454

Teräsvirta T, van Dijk D, Medeiros MC (2005) Linear models, smooth transition autoregressions, and neural networks for forecasting macroeconomic time series: a re-examination. Int J Forecast 21(3):755–774

Tsaur R, Kuo T (2011) The adaptive fuzzy time series model with an application to Taiwan's tourism demand. Expert Syst Appl 38(7):9164–9171

Weron R (2014) Electricity price forecasting: a review of the state-of-the-art with a look into the future. Int J Forecast 30(3):1030–1081

Williams CKI, Rasmussen CE (2006) Gaussian processes for machine learning. The MIT Press, Cambridge

Witt SF, Witt CA (1992) Modelling and forecasting demand in tourism. London Academic Press, London

WTTC (2016) WTTC travel and tourism economic impact 2016. World Travel and Tourism Council, Spain

Wu Q, Law R, Xu X (2012) A spare Gaussian process regression model for tourism demand forecasting in Hong Kong. Expert Syst Appl 39(4):4769–4774

Yu G, Schwartz Z (2006) Forecasting short time-series tourism demand with artificial intelligence models. J Travel Res 45(1):194–203

Zhang G, Putuwo BE, Hu MY (1998) Forecasting with artificial neural networks: the state of the art. Int J Forecast 14(1):35–62

Zhang GP, Qi M (2005) Neural network forecasting for seasonal and trend time series. Eur J Oper Res 160(1):501–514